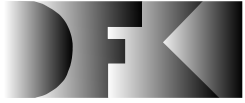

# A Complete and Recursive Feature Theory

**Rolf Backofen and Gert Smolka**

September 1992

# Deutsches Forschungszentrum für Künstliche Intelligenz

The German Research Center for Artificial Intelligence (Deutsches Forschungszentrum für Künstliche Intelligenz, DFKI) with sites in Kaiserslautern and Saarbrücken is a non-profit organization which was founded in 1988. The shareholder companies are Atlas Elektronik, Daimler-Benz, Fraunhofer Gesellschaft, GMD, IBM, Insiders, Mannesmann-Kienzle, Sema Group, Siemens and Siemens-Nixdorf. Research projects conducted at the DFKI are funded by the German Ministry for Research and Technology, by the shareholder companies, or by other industrial contracts.

The DFKI conducts application-oriented basic research in the field of artificial intelligence and other related subfields of computer science. The overall goal is to construct systems with technical knowledge and common sense which - by using AI methods - implement a problem solution for a selected application area. Currently, there are the following research areas at the DFKI:

- ☐ Intelligent Engineering Systems
- ☐ Intelligent User Interfaces
- ☐ Computer Linguistics
- ☐ Programming Systems
- ☐ Deduction and Multiagent Systems
- ☐ Document Analysis and Office Automation.

The DFKI strives at making its research results available to the scientific community. There exist many contacts to domestic and foreign research institutions, both in academy and industry. The DFKI hosts technology transfer workshops for shareholders and other interested groups in order to inform about the current state of research.

From its beginning, the DFKI has provided an attractive working environment for AI researchers from Germany and from all over the world. The goal is to have a staff of about 100 researchers at the end of the building-up phase.

Friedrich J. Wendl
Director

# A Complete and Recursive Feature Theory

**Rolf Backofen and Gert Smolka**

A short version of this report has appeared in the *Proceedings of the 31st Annual Meeting of the Association for Computational Linguistics*

This work has been supported by a grant from The Federal Ministry for Research and Technology (FKZ ITWM-9105 and ITWM-9002 0).

# A Complete and Recursive Feature Theory

Rolf Backofen and Gert Smolka

German Research Center for Artificial Intelligence (DFKI)
Stuhlsatzenhausweg 3, D-6600 Saarbrücken, Germany
{backofen, smolka}@dfki.uni-sb.de

**Abstract**

Various feature descriptions are being employed in logic programming languages and constrained-based grammar formalisms. The common notational primitive of these descriptions are functional attributes called features. The descriptions considered in this paper are the possibly quantified first-order formulae obtained from a signature of binary and unary predicates called features and sorts, respectively. We establish a first-order theory $FT$ by means of three axiom schemes, show its completeness, and construct three elementarily equivalent models.

One of the models consists of so-called feature graphs, a data structure common in computational linguistics. The other two models consist of so-called feature trees, a record-like data structure generalizing the trees corresponding to first-order terms.

Our completeness proof exhibits a terminating simplification system deciding validity and satisfiability of possibly quantified feature descriptions.

# Contents

# 1   Introduction

Feature descriptions provide for the typically partial description of abstract objects by means of functional attributes called features. They originated in the late seventies with so-called unification grammars [15, 13], a by now popular family of declarative grammar formalisms for the description and processing of natural language. More recently, the use of feature descriptions in logic programming has been advocated and studied [3, 4, 5, 6, 21]. Essentially, feature descriptions provide a logical version of records, a data structure found in many programming languages.

Feature descriptions have been proposed in various forms with various formalizations [1, 2, 14, 18, 11, 20, 7, 12]. We will follow the logical approach pioneered by [20], which accommodates feature descriptions as standard first-order formulae interpreted in first-order structures. In this approach, a semantics for feature descriptions can be given by means of a feature theory (i.e., a set of closed feature descriptions having at least one model). There are two complementary ways of specifying a feature theory: either by explicitly constructing a standard model and taking all sentences valid in it, or by stating axioms and proving their consistency. Both possibilities are exemplified in [20]: the feature graph algebra $\mathcal{F}$ is given as a standard model, and the class of feature algebras is obtained by means of an axiomatization.

Both approaches to fixing a feature theory have their advantages. The construction of a standard model provides for a clear intuition and yields a complete feature theory (i.e., if $\phi$ is a closed feature description, then either $\phi$ or $\neg\phi$ is valid). The presentation of a recursively enumerable axiomatization has the advantage that we inherit from predicate logic a sound and complete deduction system for valid feature descriptions.

The ideal case then is to specify a feature theory by both a standard model and a corresponding recursively enumerable axiomatization. The existence of such a double characterization, however, is by no means obvious since it implies that the feature theory is decidable. In fact, so far no decidable, consistent and complete feature theory has been known.

In this paper we will establish a complete and decidable feature theory $FT$ by means of three axiom schemes. We will also construct three models of $FT$, two consisting of so-called feature trees, and one consisting of so-called feature graphs. Since $FT$ is complete, all three models are elementarily equivalent (i.e., satisfy exactly the same first-order formulae). While the feature graph model captures intuitions common in linguistically motivated investigations, the feature tree model provides the connection to the tree constraint systems [9, 10, 16, 17] employed in logic programming.

Our proof of $FT$'s completeness will exhibit a simplification algorithm that computes for every feature description an equivalent solved form from which the solutions of the description can be read of easily. For a closed feature description the solved form is either $\top$ (which means that the description is valid) or $\bot$ (which means that the description is invalid). For a feature description with free variables the solved form is $\bot$ if and only if the description is unsatisfiable.

## 1.1 Feature Descriptions

Feature descriptions are first-order formulae built over an alphabet of binary predicate symbols, called *features*, and an alphabet of unary predicate symbols, called *sorts*. There are no function symbols. In admissible interpretations features must be functional relations, and distinct sorts must be disjoint sets. This is stated by the first and second axiom scheme of $FT$:

(Ax1)    $\forall x \forall y \forall z (f(x,y) \land f(x,z) \rightarrow y \doteq z)$    (for every feature $f$)

(Ax2)    $\forall x (A(x) \land B(x) \rightarrow \bot)$    (for every two distinct sorts $A$ and $B$).

A typical feature description written in matrix notation is

$$x \;:\; \exists y \begin{bmatrix} woman \\ father \;:\; \begin{bmatrix} engineer \\ age : y \end{bmatrix} \\ husband \;:\; \begin{bmatrix} painter \\ age : y \end{bmatrix} \end{bmatrix}.$$

It may be read as saying that $x$ is a woman whose father is an engineer, whose husband is a painter, and whose father and husband are both of the same age. Written in plain first-order syntax we obtain the less suggestive formula

$$\exists y, F, H \;(\; woman(X) \land$$
$$father(x, F) \land engineer(F) \land age(F, y) \land$$
$$husband(x, H) \land painter(H) \land age(H, y) \;).$$

The axiom schemes (Ax1) and (Ax2) still admit trivial models where all features and sorts are empty. The third and final axiom scheme of $FT$ states that certain "consistent" descriptions have solutions. Three Examples of instances of $FT$'s third axiom scheme are

$$\exists x, y, z \;(f(x,y) \land A(y) \land g(x,z) \land B(z))$$
$$\forall u, z \;\exists x, y \;(f(x,y) \land g(y,u) \land h(y,z) \land yf\!\uparrow)$$
$$\forall z \;\exists x, y \;(f(x,y) \land g(y,x) \land h(y,z) \land yf\!\uparrow),$$

where $yf\uparrow$ abbreviates $\neg\exists z(f(y,z))$. Note that the third description

$$f(x,y) \wedge g(y,x) \wedge h(y,z) \wedge fy\uparrow$$

is "cyclic" with respect to the variables $x$ and $y$.

## 1.2 Feature Trees

A feature tree (examples are shown in Figure 1) is a tree whose edges are labeled with features, and whose nodes are labeled with sorts. As one would expect, the labeling with features must be deterministic, that is, the direct subtrees of a feature tree must be uniquely identified by the features of the edges leading to them. Feature trees can be seen as a mathematical model of records in programming languages. Feature trees without subtrees model atomic values (e.g., numbers). Feature trees may be finite or infinite, where infinite feature trees provide for the convenient representation of cyclic data structures. The last example in Figure 1 gives a finite graph representation of an infinite feature tree, which may arise as the representation of the recursive type equation $nat = 0 + s(nat)$.

A ground term, say $f(g(a,b), h(c))$, can be seen as a feature tree whose nodes are labeled with function symbols and whose arcs are labeled with numbers:

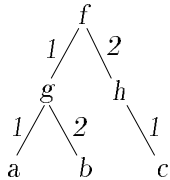

Thus the trees corresponding to first-order terms are in fact feature trees observing certain restrictions (e.g., the features departing from a node must be consecutive positive integers).

Feature descriptions are interpreted over feature trees as one would expect:

- Every sort symbol $A$ is taken as a unary predicate, where a *sort constraint* $A(x)$ holds if and only if the root of the tree $x$ is labeled with $A$.

- Every feature symbol $f$ is taken as a binary predicate, where a *feature constraint* $f(x,y)$ holds if and only if the tree $x$ has the direct subtree $y$ at feature $f$.

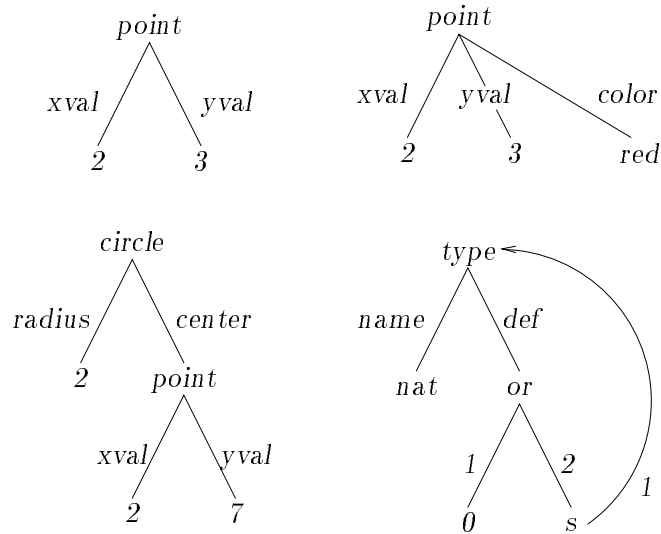

Figure 1: Examples of Feature Trees.

The theory of the corresponding first-order structure (i.e., the set of all closed formulae valid in this structure) is called $FT$. We will show that $FT$ is in fact exactly the theory specified by the three axiom schemes outlined above, provided the alphabets of sorts and features are both taken to be infinite. Hence $FT$ is complete (since it is the theory of the feature tree structure) and decidable (since it is complete and specified by a recursive set of axioms).

Another, elementarily equivalent, model of $FT$ is the substructure of the feature tree structure obtained by admitting only rational feature trees (i.e., finitely branching trees having only finitely many subtrees). Yet another model of $FT$ can be obtained from so-called feature graphs, which are finite, directed, possibly cyclic graphs labelled with sorts and features similar to feature trees. In contrast to feature trees, nodes of feature graphs may or may not be labelled with sorts. Feature graphs correspond to the so-called feature structures commonly found in linguistically motivated investigations [19, 8].

### 1.3  Organization of the Paper

Section 2 recalls the necessary notions and notations from Predicate Logic. Section 3 defines the theory $FT$ by means of three axiom schemes. Section 4 establishes the overall structure of the completeness proof by means of a lemma. Section 5 studies quantifier-free conjunctive formulae, gives a solved form, and introduces path constraints. Section 6 defines feature trees and graphs and establishes the respective models of *FT*. Section 7 studies the properties of so-called prime formulae, which are the basic building stones of the solved form for general feature constraints. Section 8 presents the quantifier elimination lemmas and completes the completeness proof.

## 2 Preliminaries

Throughout this paper we assume a signature SOR ⊎ FEA consisting of an infinite set SOR of unary predicate symbols called **sorts** and an infinite set FEA of binary predicate symbols called **features**. For the completeness of our axiomatization it is essential that there are both infinitely many sorts and infinitely many features.[1] The letters $A$, $B$, $C$ will always denote sorts, and the letters $f$, $g$, $h$ will always denote features.

A **path** is a word (i.e., a finite, possibly empty sequence) over the set of all features. The symbol $\varepsilon$ denotes the empty path, which satisfies $\varepsilon p = p = p\varepsilon$ for every path $p$. A path $p$ is called a **prefix** of a path $q$, if there exists a path $p'$ such that $pp' = q$.

We also assume an infinite alphabet of variables and adopt the convention that $x$, $y$, $z$ always denote variables, and $X$, $Y$ always denote finite, possibly empty sets of variables. Under our signature SOR ⊎ FEA, every term is a variable, and an atomic formula is either a **feature constraint** $xfy$ ($f(x,y)$ in standard notation), a **sort constraint** $Ax$ ($A(x)$ in standard notation), an equation $x \doteq y$, $\bot$ ("false"), or $\top$ ("true"). Compound formulae are obtained as usual with the connectives $\wedge$, $\vee$, $\rightarrow$, $\leftrightarrow$, $\neg$ and the quantifiers $\exists$ and $\forall$. We use $\tilde{\exists}\phi$ [$\tilde{\forall}\phi$] to denote the existential [universal] closure of a formula $\phi$. Moreover, $\mathcal{V}(\phi)$ is taken to denote the set of all variables that occur free in a formula $\phi$. The letters $\phi$ and $\psi$ will always denote formulae.

We assume that the conjunction of formulae is an associative and commutative operation that has $\top$ as neutral element. This means that we identify $\phi \wedge (\psi \wedge \theta)$ with $\theta \wedge (\psi \wedge \phi)$, and $\phi \wedge \top$ with $\phi$ (but not, for example, $xfy \wedge xfy$ with $xfy$). A conjunction of atomic formulae can thus be seen as the finite multiset of these formulae, where conjunction is multiset union, and $\top$ (the "empty conjunction") is the empty multiset. We will write $\psi \subseteq \phi$ (or $\psi \in \phi$, if $\psi$ is an atomic formula) if there exists a formula $\psi'$ such that $\psi \wedge \psi' = \phi$.

Moreover, we identify $\exists x \exists y \phi$ with $\exists y \exists x \phi$. If $X = \{x_1, \ldots, x_n\}$, we write $\exists X \phi$ for $\exists x_1 \ldots \exists x_n \phi$. If $X = \emptyset$, then $\exists X \phi$ stands for $\phi$.

---
[1]The assumption that the alphabets of sorts and features are infinite is used in Proposition 7.9 and Lemma 8.4.

Structures and satisfaction of formulae are defined as usual. A valuation into a structure $\mathcal{A}$ is a total function from the set of all variables into the universe $|\mathcal{A}|$ of $\mathcal{A}$. A valuation $\alpha'$ into $\mathcal{A}$ is called an $x$-**update** [$X$-**update**] of a valuation $\alpha$ into $\mathcal{A}$ if $\alpha'$ and $\alpha$ agree everywhere but possibly on $x$ [$X$]. We use $\phi^{\mathcal{A}}$ to denote the set of all valuations $\alpha$ such that $\mathcal{A}, \alpha \models \phi$. We write $\phi \models \psi$ ("$\phi$ entails $\psi$") if $\phi^{\mathcal{A}} \subseteq \psi^{\mathcal{A}}$ for all structures $\mathcal{A}$, and $\phi \models\mid \psi$ ("$\phi$ is equivalent to $\psi$") if $\phi^{\mathcal{A}} = \psi^{\mathcal{A}}$ for all structures $\mathcal{A}$.

A **theory** is a set of closed formulae. A **model** of a theory is a structure that satisfies every formulae of the theory. A formula $\phi$ is a **consequence of a theory** $T$ ($T \models \phi$) if $\tilde{\forall}\phi$ is valid in every model of $T$. A formula $\phi$ **entails** a formula $\psi$ in a theory $T$ ($\phi \models_T \psi$) if $\phi^{\mathcal{A}} \subseteq \psi^{\mathcal{A}}$ for every model $\mathcal{A}$ of $T$. Two formulae $\phi, \psi$ are **equivalent** in a theory $T$ ($\phi \models\mid_T \psi$) if $\phi^{\mathcal{A}} = \psi^{\mathcal{A}}$ for every model $\mathcal{A}$ of $T$.

A theory $T$ is **complete** if for every closed formula $\phi$ either $\phi$ or $\neg\phi$ is a consequence of $T$. A theory is **decidable** if the set of its consequences is decidable. Since the consequences of a recursively enumerable theory are recursively enumerable (completeness of first-order deduction), a complete theory is decidable if and only if it is recursively enumerable.

Two first-order structures $\mathcal{A}, \mathcal{B}$ are **elementarily equivalent** if, for every first-order formula $\phi$, $\phi$ is valid in $\mathcal{A}$ if and only if $\phi$ is valid in $\mathcal{B}$. Note that all models of a complete theory are elementarily equivalent.

## 3 The Axioms

The first axiom scheme says that features are functional:

*(Ax1)* $\quad \forall x \forall y \forall z (xfy \wedge xfz \to y \doteq z) \quad$ (for every feature $f$).

The second scheme says that sorts are mutually disjoint:

*(Ax2)* $\quad \forall x (Ax \wedge Bx \to \bot) \quad$ (for every two distinct sorts $A$ and $B$).

The third and final axiom scheme will say that certain "consistent feature descriptions" are satisfiable. For its formulation we need the important notion of a solved clause.

An **exclusion constraint** is an additional atomic formula of the form $xf\uparrow$ ("$f$ undefined on $x$") taken to be equivalent to $\neg \exists y\, (xfy)$ (for some variable $y \neq x$).

A **solved clause** is a possibly empty conjunction $\phi$ of atomic formulae of the form $xfy$, $Ax$ and $xf\uparrow$ such that the following conditions are satisfied:

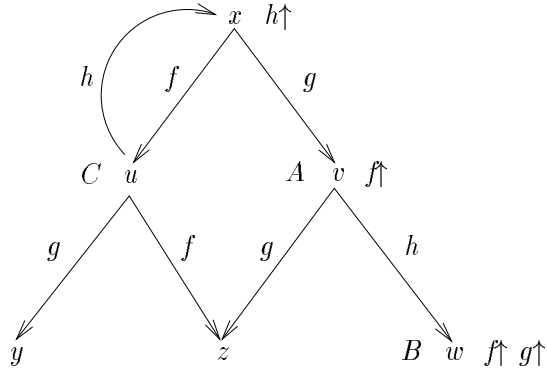

Figure 2: A graph representation of a solved clause.

1. no atomic formula occurs twice in $\phi$

2. if $Ax \in \phi$ and $Bx \in \phi$, then $A = B$

3. if $xfy \in \phi$ and $xfz \in \phi$, then $y = z$

4. if $xfy \in \phi$, then $xf\uparrow \notin \phi$.

Figure 2 gives a graph representation of the solved clause

$$xfu \wedge xgv \wedge xh\uparrow \wedge$$
$$Cu \wedge uhx \wedge ugy \wedge ufz \wedge$$
$$Av \wedge vgz \wedge vhw \wedge vf\uparrow \wedge$$
$$Bw \wedge wf\uparrow \wedge wg\uparrow \;.$$

A more readable textual representation of this solved clause is

$$\begin{array}{rcl} x & : & [f{:}u \;\; g{:}v \;\; h\uparrow] \\ u & : & [C \;\; h{:}x \;\; g{:}y \;\; f{:}z] \\ v & : & [A \;\; g{:}z \;\; h{:}w \;\; f\uparrow] \\ w & : & [B \;\; f\uparrow \;\; g\uparrow]. \end{array}$$

As in the example, a solved clause can always be seen as the graph whose nodes are the variables appearing in the clause and whose arcs are given by the feature constraints $xfy$. The constraints $Ax$, $xf\uparrow$ appear as labels of the node $x$. The graphical representation of solved clauses should be very helpful in understanding the proofs to come.

A variable $x$ is **constrained** in a solved clause $\phi$ if $\phi$ contains a constraint of the form $Ax$, $xfy$ or $xf\uparrow$. We use $\mathcal{CV}(\phi)$ to denote the set of all variables that are constrained in $\phi$. The variables in $\mathcal{V}(\phi) - \mathcal{CV}(\phi)$ are called the **parameters** of a solved clause $\phi$. In the graph representation of a solved clause the parameters appear as leaves that are not not labeled with a sort or a feature exclusion. The parameters of the solved clause in Figure 2 are $y$ and $z$.

We can now state the third axiom scheme. It says that the constrained variables of a solved clause have solutions for all values of the parameters:

(Ax3)  $\tilde{\forall} \exists X \phi$  (for every solved clause $\phi$ and $X = \mathcal{CV}(\phi)$).

The **theory** $FT$ is the set of all sentences that can be obtained as instances of the axiom schemes (Ax1), (Ax2) and (Ax3). The **theory** $FT_0$ is the set of all sentences that can be obtained as instances of the first two axiom schemes.

As the main result of this paper we will show that $FT$ is a complete and decidable theory.

By using an adaption of the proof of Theorem 8.3 in [20] one can show that $FT_0$ is undecidable.

## 4 Outline of the Completeness Proof

The completeness of $FT$ will be shown by exhibiting a simplification algorithm for $FT$. The following lemma gives the overall structure of the algorithm, which is the same as in Maher's [17] completeness proof for the theory of constructor trees.

**Lemma 4.1** *Suppose there exists a set of so-called prime formulae such that:*

1. *every sort constraint $Ax$, every feature constraint $xfy$, and every equation $x \doteq y$ such that $x \neq y$ is a prime formula*

2. *$\top$ is a prime formula, and there is no other closed prime formula*

3. *for every two prime formulae $\beta$ and $\beta'$ one can compute a formula $\delta$ that is either prime or $\bot$ and satisfies*

$$\beta \wedge \beta' \models_{FT} \delta \quad \text{and} \quad \mathcal{V}(\delta) \subseteq \mathcal{V}(\beta \wedge \beta')$$

4. *for every prime formula $\beta$ and every variable $x$ one can compute a prime formula $\beta'$ such that*

$$\exists x \beta \models_{FT} \beta' \quad \text{and} \quad \mathcal{V}(\beta') \subseteq \mathcal{V}(\exists x \beta)$$

5. if $\beta$, $\beta_1, \ldots, \beta_n$ are prime formulae, then

$$\exists x (\beta \wedge \bigwedge_{i=1}^{n} \neg \beta_i) \models_{FT} \bigwedge_{i=1}^{n} \exists x (\beta \wedge \neg \beta_i)$$

6. for every two prime formulae $\beta$, $\beta'$ and every variable $x$ one can compute a Boolean combination $\delta$ of prime formulae such that

$$\exists x (\beta \wedge \neg \beta') \models_{FT} \delta \quad \text{and} \quad \mathcal{V}(\delta) \subseteq \mathcal{V}(\exists x (\beta \wedge \neg \beta')).$$

Then one can compute for every formula $\phi$ a Boolean combination $\delta$ of prime formulae such that $\phi \models_{FT} \delta$ and $\mathcal{V}(\delta) \subseteq \mathcal{V}(\phi)$.

**Proof.** Suppose a set of prime formulae as required exists. Let $\phi$ be a formula. We show by induction on the structure of $\phi$ how to compute a Boolean combination $\delta$ of prime formulae such that $\phi \models_{FT} \delta$ and $\mathcal{V}(\delta) \subseteq \mathcal{V}(\phi)$.

If $\phi$ is an atomic formula $Ax$, $xfy$ or $x \doteq y$, then $\phi$ is either a prime formula, or $\phi$ is a trivial equation $x \doteq x$, in which case it is equivalent to the prime formula $\top$.

If $\phi$ is $\neg \psi$, $\psi \wedge \psi'$ or $\psi \vee \psi'$, then the claim follows immediately with the induction hypothesis.

It remains to show the claim for $\phi = \exists x \psi$. By the induction hypothesis we know that we can compute a Boolean combination $\delta$ of prime formulae such that $\delta \models_{FT} \psi$ and $\mathcal{V}(\delta) \subseteq \mathcal{V}(\psi)$. Now $\delta$ can be transformed to a disjunctive normal form where prime formulae play the role of atomic formulae; that is, $\delta$ is equivalent to $\sigma_1 \vee \ldots \vee \sigma_n$, where every "clause" $\sigma_i$ is a conjunction of prime and negated prime formulae. Hence

$$\exists x \phi \models \exists x (\sigma_1 \vee \ldots \vee \sigma_n) \models \exists x \sigma_1 \vee \ldots \vee \exists x \sigma_n,$$

where all three formulae have exactly the same free variables. It remains to show that one can compute for every clause $\sigma$ a Boolean combination $\delta$ of prime formulae such that $\exists x \sigma \models_{FT} \delta$ and $\mathcal{V}(\delta) \subseteq \mathcal{V}(\exists x \sigma)$. We distinguish the following cases.
(i) $\sigma = \beta$ for some basic formula $\beta$. Then the claim follows by assumption (4).
(ii) $\sigma = \beta \wedge \bigwedge_{i=1}^{n} \neg \beta_i$, $n > 0$. Then the claim follows with assumptions (5) and (6).
(iii) $\sigma = \bigwedge_{i=1}^{n} \neg \beta_i$, $n > 0$. Then $\sigma \models_{FT} \top \wedge \bigwedge_{i=1}^{n} \neg \beta_i$ and the claim follows with case (ii) since $\top$ is a prime formula by assumption (2).
(iv) $\sigma = \beta_1 \wedge \ldots \wedge \beta_k \wedge \neg \beta'_1 \wedge \ldots \wedge \neg \beta'_n$, $k > 1$, $n \geq 0$. Then we know by

assumption (3) that either $\beta_1 \wedge \ldots \wedge \beta_k \models_{FT} \bot$ or $\beta_1 \wedge \ldots \wedge \beta_k \models_{FT} \beta$ for some prime formula $\beta$. In the former case we choose $\delta = \neg\top$, and in the latter case the claim follows with case (i) or (ii). □

Note that, provided a set of prime formulae with the required properties exists, the preceding lemma yields the completeness of $FT$ since every closed formula can be simplified to $\top$ or $\neg\top$ (since $\top$ is the only closed prime formula).

In the following we will establish a set of prime formula as required.

## 5 Solved Formulae

In this section we introduce a solved form for conjunctions of atomic formulae.

A **basic formula** is either $\bot$ or a possibly empty conjunction of atomic formulae of the form $Ax$, $xfy$, and $x \doteq y$. Note that $\top$ is a basic formula since $\top$ is the empty conjunction.

Every basic formula $\phi \neq \bot$ has a unique decomposition $\phi = \phi_N \wedge \phi_G$ into a possibly empty conjunction $\phi_N$ of equations "$x \doteq y$" and a possibly empty conjunction $\phi_G$ of sort constraints "$Ax$" and feature constraints "$xfy$". We call $\phi_N$ the **normalizer** and and $\phi_G$ the **graph** of $\phi$.

We say that a basic formula $\phi$ **binds** $x$ **to** $y$ if $x \doteq y \in \phi$ and $x$ occurs only once in $\phi$. Here it is important to note that we consider equations as directed, that is, assume that $x \doteq y$ is different from $y \doteq x$ if $x \neq y$. We say that $\phi$ **eliminates** $x$ if $\phi$ binds $x$ to some variable $y$.

A **solved formula** is a basic formula $\gamma \neq \bot$ such that the following conditions are satisfied:

1. an equation $x \doteq y$ appears in $\gamma$ if and only if $\gamma$ eliminates $x$

2. the graph of $\gamma$ is a solved clause.

Note that a solved clause not containing exclusion constraints is a solved formula, and that a solved formula not containing equations is a solved clause. The letter $\gamma$ will always denote a solved formula.

We will see that every basic formula is equivalent in $FT_0$ to either $\bot$ or a solved formula.

1. $$\frac{xfy \wedge xfz \wedge \phi}{xfz \wedge y \doteq z \wedge \phi}$$

2. $$\frac{Ax \wedge Bx \wedge \phi}{\bot} \quad A \neq B$$

3. $$\frac{Ax \wedge Ax \wedge \phi}{Ax \wedge \phi}$$

4. $$\frac{x \doteq y \wedge \phi}{x \doteq y \wedge \phi[x \leftarrow y]} \quad x \in \mathcal{V}(\phi) \text{ and } x \neq y$$

5. $$\frac{x \doteq x \wedge \phi}{\phi}$$

Figure 3: The basic simplification rules.

Figure 3 shows the so-called **basic simplification rules**. With $\phi[x \leftarrow y]$ we denote the formula that is obtained from $\phi$ by replacing every occurrence of $x$ with $y$. We say that a formula $\phi$ **simplifies to** a formula $\psi$ by a simplification rule $\rho$ if $\frac{\phi}{\psi}$ is an instance of $\rho$. We say that a basic formula $\phi$ **simplifies to** a basic formula $\psi$ if either $\phi = \psi$ or $\phi$ simplifies to $\psi$ in finitely many steps each licensed by one of basic simplification rules in Figure 3.

Note that the basic simplification rules (1) and (2) correspond to the first and second axiom scheme, respectively. Thus they are equivalence transformation with respect to $FT_0$. The remaining three simplification rules are equivalence transformations in general.

**Proposition 5.1** *The basic simplification rules are terminating and perform equivalence transformations with respect to $FT_0$. Moreover, a basic formula $\phi \neq \bot$ is solved if and only if no basic simplification rule applies to it.*

**Proof.** To see that the basic simplification rules are terminating, observe that no rule adds a new variable and that every rule preserves eliminated variables. Since rule (4) increases the number of eliminated variables, and the remaining rules obviously terminate, the entire system must terminate. The other claims are easy to verify. □

**Proposition 5.2** *Let $\phi$ be a formula built from atomic formulae with conjunction. Then one can compute a formula $\delta$ that is either solved or $\bot$ such* that $\phi \mathrel{\mathop{\vDash}\limits^{\mid}}_{FT_0} \delta$ and $\mathcal{V}(\delta) \subseteq \mathcal{V}(\phi)$.

**Proof.** Follows from the preceding proposition and the fact that the basic simplification rules do not introduce new variables. $\square$

In the quantifier elimination proofs to come it will be convenient to use so-called path constraints, which provide a flexible syntax for atomic formulae closed under conjunction and existential quantification. We start by defining the denotation of a path.

The interpretations $f^{\mathcal{A}}$, $g^{\mathcal{A}}$ of two features $f$, $g$ in a structure $\mathcal{A}$ are binary relations on the universe $|\mathcal{A}|$ of $\mathcal{A}$; hence their composition $f^{\mathcal{A}} \circ g^{\mathcal{A}}$ is again a binary relation on $|\mathcal{A}|$ satisfying

$$a(f^{\mathcal{A}} \circ g^{\mathcal{A}})b \iff \exists c \in |\mathcal{A}|: \ af^{\mathcal{A}}c \ \wedge \ cf^{\mathcal{A}}b$$

for all $a, b \in |\mathcal{A}|$. Consequently we define the **denotation $p^{\mathcal{A}}$ of a path** $p = f_1 \cdots f_n$ in a structure $\mathcal{A}$ as the composition

$$(f_1 \cdots f_n)^{\mathcal{A}} := f_1^{\mathcal{A}} \circ \cdots \circ f_n^{\mathcal{A}},$$

where the empty path $\varepsilon$ is taken to denote the identity relation. If $\mathcal{A}$ is a model of the theory $FT_0$, then every paths denotes a unary partial function on the universe of $\mathcal{A}$. Given an element $a \in |\mathcal{A}|$, $p^{\mathcal{A}}$ is thus either undefined on $a$ or leads from $a$ to exactly one $b \in |\mathcal{A}|$.

Let $p$, $q$ be paths, $x$, $y$ be variables, and $A$ be a sort. Then **path constraints** are defined as follows:

$$\begin{aligned} \mathcal{A}, \alpha &\models xpy &:&\iff \alpha(x) \, p^{\mathcal{A}} \, \alpha(y) \\ \mathcal{A}, \alpha &\models xp{\downarrow}yq &:&\iff \exists a \in |\mathcal{A}|: \ \alpha(x) \, p^{\mathcal{A}} \, a \ \wedge \ \alpha(y) \, q^{\mathcal{A}} \, a \\ \mathcal{A}, \alpha &\models Axp &:&\iff \exists a \in |\mathcal{A}|: \ \alpha(x) \, p^{\mathcal{A}} \, a \ \wedge \ a \in A^{\mathcal{A}}. \end{aligned}$$

Note that path constraints $xpy$ generalize feature constraints $xfy$.

A **proper path constraint** is a path constraint of the form "$Axp$" or "$xp{\downarrow}yq$".

Every path constraint can be expressed with the already existing formulae, as can be seen from the following equivalences:

$$\begin{aligned} x\varepsilon y &\mathrel{\mathop{\vDash}\limits^{\mid}} x \doteq y \\ xfpy &\mathrel{\mathop{\vDash}\limits^{\mid}} \exists z(xfz \wedge zpy) &(z \neq x, y) \\ xp{\downarrow}yq &\mathrel{\mathop{\vDash}\limits^{\mid}} \exists z(xpz \wedge yqz) &(z \neq x, y) \\ Axp &\mathrel{\mathop{\vDash}\limits^{\mid}} \exists y(xpy \wedge Ay) &(y \neq x). \end{aligned}$$

The **closure** $[\gamma]$ **of a solved formula** $\gamma$ is the closure of the atomic formulae occurring in $\gamma$ with respect to the following deduction rules:

$$\frac{}{x\varepsilon x} \qquad \frac{x \doteq y}{x\varepsilon y} \qquad \frac{xpy \quad yfz}{xpfz} \qquad \frac{xpz \quad yqz}{xp{\downarrow}yq} \qquad \frac{Ay \quad xpy}{Axp}.$$

Recall that we assume that equations $x \doteq y$ are directed, that is, are ordered pairs of variables. Hence, $x\varepsilon y \in [\gamma]$ and $y\varepsilon x \notin [\gamma]$ if $x \doteq y \in \gamma$.

The **closure of a solved clause** $\delta$ is defined analogously.

**Proposition 5.3** *Let $\gamma$ be a solved formula. Then:*

1. *if $\pi \in [\gamma]$, then $\gamma \models \pi$*
2. *$x\varepsilon y \in [\gamma]$ iff $x = y$ or $x \doteq y \in \gamma$*
3. *$xfy \in [\gamma]$ iff $xfy \in \gamma$ or $\exists z$: $x \doteq z \in \gamma$ and $zfy \in \gamma$*
4. *$xpfy \in [\gamma]$ iff $\exists z$: $xpz \in [\gamma]$ and $zfy \in \gamma$*
5. *if $p \neq \varepsilon$ and $xpy, xpz \in [\gamma]$, then $y = z$*
6. *it is decidable whether a path constraint is in $[\gamma]$.*

**Proof.** For the first claim one verifies the soundness of the deduction rules for path constraints. The verification of the other claims is straightforward. □

## 6 Feature Trees and Feature Graphs

In this section we establish three models of $FT$ consisting of either feature trees or feature graphs. Since we will show that $FT$ is a complete theory, all three models are in fact elementarily equivalent.

A **tree domain** is a nonempty set $D \subseteq \text{FEA}^\star$ of paths that is **prefix-closed**, that is, if $pq \in D$, then $p \in D$. Note that every tree domain contains the empty path.

A **feature tree** is a partial function $\sigma: \text{FEA}^\star \to \text{SOR}$ whose domain is a tree domain. The paths in the domain of a feature tree represent the nodes of the tree; the empty path represents its root. We use $D_\sigma$ to denote the domain of a feature tree $\sigma$. A feature tree is called **finite** [**infinite**] if its domain is finite [infinite]. The letters $\sigma$ and $\tau$ will always denote feature trees.

The **subtree** $p\sigma$ of a feature tree $\sigma$ at a path $p \in D_\sigma$ is the feature tree defined by (in relational notation)

$$p\sigma \;:=\; \{(q, A) \mid (pq, A) \in \sigma\}.$$

A feature tree $\sigma$ is called a **subtree** of a feature tree $\tau$ if $\sigma$ is a subtree of $\tau$ at some path $p \in D_\tau$, and a **direct subtree** if $p = f$ for some feature $f$.

A feature tree $\sigma$ is called **rational** if (1) $\sigma$ has only finitely many subtrees and (2) $\sigma$ is finitely branching (i.e., for every $p \in D_\sigma$, the set $\{pf \in D_\sigma \mid f \in \mathrm{FEA}\}$ is finite). Note that for every rational feature tree $\sigma$ there exist finitely many features $f_1, \ldots, f_n$ such that $D_\sigma \subseteq \{f_1, \ldots, f_n\}^\star$.

The **feature tree structure** $\mathcal{T}$ is the SOR $\uplus$ FEA-structure defined as follows:

- the universe of $\mathcal{T}$ is the set of all feature trees
- $\sigma \in A^\mathcal{T}$ iff $\sigma(\varepsilon) = A$ (i.e., $\sigma$'s root is labeled with $A$)
- $(\sigma, \tau) \in f^\mathcal{T}$ iff $f \in D_\sigma$ and $\tau = f\sigma$ (i.e., $\tau$ is the subtree of $\sigma$ at $f$).

The **rational feature tree structure** $\mathcal{R}$ is the substructure of $\mathcal{T}$ consisting only of the rational feature trees.

**Theorem 6.1** *The feature tree structures $\mathcal{T}$ and $\mathcal{R}$ are models of the theory FT.*

**Proof.** We will first show that $\mathcal{T}$ is a model of *FT*.

The first and second axiom scheme are obviously satisfied by $\mathcal{T}$. To see that $\mathcal{T}$ satisfies the third axiom scheme, let $\delta$ be a solved clause, $X$ be the variables constrained in $\delta$, and $\alpha$ be a valuation into $\mathcal{T}$. It suffices to show that there exists an $X$-update $\alpha'$ of $\alpha$ such that $\mathcal{T}, \alpha' \models \delta$.

Without loss of generality we can assume that $\delta$ contains a sort constraint $Ax$ for every $x \in X$. Now one can verify that

$$\forall x \in X\colon$$
$$(p, A) \in \alpha'(x) \iff Axp \in [\delta] \;\vee$$
$$\exists xp'y \in [\delta]\; \exists (p'', A) \in \alpha(y)\colon\; p = p'p'' \wedge y \notin X$$

defines an $X$-update $\alpha'$ of $\alpha$ such that $\mathcal{T}, \alpha' \models \delta$.

The same construction shows that $\mathcal{R}$ is a model of *FT*. $\square$

A **feature pregraph** is a pair $(x, \gamma)$ consisting of a variable $x$ (called the **root**) and a solved clause $\gamma$ not containing exclusion constraints such that, for every variable $y$ occurring in $\gamma$, there exists a path $p$ satisfying $xpy \in [\gamma]$. If one deletes the exclusion constraints in Figure 2, one obtains the graphical representation of a feature pregraph whose root is $x$.

A feature pregraph $(x, \gamma)$ is called a **subpregraph** of a feature pregraph $(y, \delta)$ if $\gamma \subseteq \delta$ and $x = y$ or $x \in \mathcal{V}(\delta)$. Note that a feature pregraph has only finitely many subpregraphs.

We say that two feature pregraphs are **equivalent** if they are equal up to consistent variable renaming. For instance, $(x, xfy \wedge ygx)$ and $(u, ufx \wedge xgu)$ are equivalent feature pregraphs.

A **feature graph** is an element of the quotient of the set of all feature pregraphs with respect to equivalence as defined above. We use $\overline{(x, \gamma)}$ to denote the feature graph obtained as the equivalence class of the feature pregraph $(x, \gamma)$.

In contrast to feature trees, not every node of a feature graph must carry a sort.

The **feature graph structure** $\mathcal{G}$ is the SOR $\uplus$ FEA-structure defined as follows:

- the universe of $\mathcal{G}$ is the set of all feature graphs
- $\overline{(x, \gamma)} \in A^{\mathcal{G}}$ iff $Ax \in \gamma$
- $(\overline{(x, \gamma)}, \sigma) \in f^{\mathcal{G}}$ iff there exists a maximal feature subpregraph $(y, \delta)$ of $(x, \gamma)$ such that $xfy \in \gamma$ and $\sigma = \overline{(y, \delta)}$.

**Theorem 6.2** *The feature graph structure $\mathcal{G}$ is a model of the theory FT.*

**Proof.** The first and second axiom scheme are obviously satisfied by $\mathcal{G}$. To see that $\mathcal{G}$ satisfies the third axiom scheme, let $\delta$ be a solved clause and $\alpha$ a valuation into $\mathcal{T}$. It suffices to show that there exists an $\mathcal{CV}(\delta)$-update $\alpha'$ of $\alpha$ such that $\mathcal{G}, \alpha' \models \delta$.

First we choose for the parameters $y \in \mathcal{V}(\delta) - \mathcal{CV}(\delta)$ variable disjoint feature pregraphs $(y, \gamma_y)$ such that $\alpha(y) = \overline{(y, \gamma_y)}$. Moreover, we can assume without loss of generality that every pregraph $(y, \gamma_y)$ has with $\delta$ exactly its root variable $y$ in common. Hence

$$\delta' := \delta \wedge \bigwedge_{y \in \mathcal{V}(\delta) - \mathcal{CV}(\delta)} \gamma_y$$

is a solved clause. Now, for every constrained variable $x \in \mathcal{CV}(\delta)$, let $\rho_x$ be the maximal solved clause such that $\rho_x \subseteq \delta'$ and $(x, \rho_x)$ is a feature pregraph. Then the $\mathcal{CV}(\delta)$-update $\alpha'$ of $\alpha$ such that $\alpha'(x) = \overline{(x, \rho_x)}$ for every $x \in \mathcal{CV}(\delta)$ satisfies $\mathcal{G}, \alpha' \models \delta$. □

Let $\mathcal{F}$ be the structure whose domain consists of all feature pregraphs and that is otherwise defined analogous to $\mathcal{G}$. Note that $\mathcal{G}$ is in fact the quotient of $\mathcal{F}$ with respect to equivalence of feature pregraphs.

**Proposition 6.3** *The feature pregraph structure $\mathcal{F}$ is a model of $FT_0$ but not of $FT$.*

**Proof.** It is easy to see that $\mathcal{F}$ satisfies the first and second axiom scheme. To see that $\mathcal{F}$ does not satisfy the third axiom scheme, consider the solved clause

$$\delta \;\;=\;\; xfy \wedge xgz$$

and a valuation $\alpha$ into $\mathcal{F}$ such that $\alpha(y) = (x, Ax)$, $\alpha(z) = (x, Bx)$, and $A \neq B$. Then there exists no $x$-update $\alpha'$ of $\alpha$ satisfying $\mathcal{F}, \alpha' \models \delta$ since a feature pregraph cannot contain both $Ax$ and $Bx$. □

# 7 Prime Formulae

We now define a class of prime formulae having the properties required by Lemma 4.1. The prime formulae will turn out to be solved forms for formulae built from atomic formulae with conjunction and existential quantification.

A **prime formula** is a formula $\exists X \gamma$ such that

1. $\gamma$ is a solved formula
2. $X$ has no variable in common with the normalizer of $\gamma$
3. every $x \in X$ can be reached from a free variable, that is, there exists a path constraint $ypx \in [\gamma]$ such that $y \notin X$.

The letter $\beta$ will always denote a prime formula.

Note that $\top$ is the only closed prime formula, and that $\exists X \gamma$ is a prime formula if $\exists x \exists X \gamma$ is a prime formula. Moreover, every solved formula is a prime formula, and every quantifier-free prime formula is a solved formula.

The definition of prime formulae certainly fulfills the requirements (1) and
(2) of Lemma 4.1. The fulfillment of the requirements (3) and (4) will be
shown in this section, and the fulfillment of the requirements (5) and (6)
will be shown in the next section.

**Proposition 7.1** *Let $\exists X\gamma$ be a prime formula, $\mathcal{A}$ be a model of FT, and
$\mathcal{A}, \alpha \models \exists X\gamma$. Then there exists one and only one $X$-update $\alpha'$ of $\alpha$ such
that $\mathcal{A}, \alpha' \models \gamma$.*

**Proof.** The existence of an $X$-update $\alpha'$ of $\alpha$ such that $\mathcal{A}, \alpha' \models \gamma$ is obvious.
The uniqueness of $\alpha'$ follows from the fact that features are functional, and
that, for every $x \in X$, there exists a "global" variable $x' \notin X$ and a path $p$
such that $\mathcal{A}, \alpha' \models x'px$ (since $x'px \in [\gamma]$). □

The next proposition establishes that prime formulae are closed under exis-
tential quantification (property (4) of Lemma 4.1). Its proof makes for the
first time use of the third axiom scheme.

**Proposition 7.2** *For every prime formula $\beta$ and every variable $x$ one can
compute a prime formula $\beta'$ such that*

$$\exists x\beta \models\!\mid_{FT} \beta' \quad \text{and} \quad \mathcal{V}(\beta') \subseteq \mathcal{V}(\exists x\beta).$$

**Proof.** Let $\beta = \exists X\gamma$ be a prime formula and $x$ be a variable. We con-
struct a prime formula $\beta'$ such that $\exists x\beta \models\!\mid_{FT} \beta'$ and $\mathcal{V}(\beta') \subseteq \mathcal{V}(\exists x\beta)$. We
distinguish the following cases.

1. $x \notin \mathcal{V}(\beta)$. Then $\beta' := \beta$ does the job.

2. $\gamma = (x \doteq y \wedge \gamma')$. Then $\beta' := \exists X\gamma'$ does the job.

3. $\gamma = (y \doteq x \wedge \gamma')$. Then $\beta' := \exists X(\gamma'[x \leftarrow y])$ does the job since $\gamma \models\!\mid x \doteq y \wedge \gamma'[x \leftarrow y]$.

4. $x \notin X$ and $x$ occurs in the graph but not in the normalizer of $\gamma$. Then
we obtain $\beta'$ by a "garbage collection" deleting all parts of $\exists x\beta$ that cannot
be reached from "global" variables. To do this we define the following:

$$\begin{aligned} Y &:= X \cup \{x\} & \text{"quantified variables"} \\ Y_1 &:= \{x \in Y \mid \exists ypx \in [\gamma]\colon y \notin Y\} & \text{"reachable variables"} \\ Y_2 &:= Y - Y_1 & \text{"unreachable variables"}. \end{aligned}$$

Furthermore, let

$$\gamma \;=\; \gamma_N \wedge \gamma_G$$

be the decomposition of $\gamma$ into normalizer and graph, and let

$$\gamma_G = \gamma'_G \wedge \gamma''_G$$

be the decomposition of $\gamma_G$ obtained by putting into $\gamma''_G$ all atomic formulae that contain a variable in $Y_2$. To stay with the garbage collection metaphor, think of $\gamma'_G$ as the reachable and of $\gamma''_G$ as the unreachable part of $\gamma_G$ (under the quantification $\exists x \exists X$).

Since $Y \subseteq \mathcal{V}(\gamma_G) - \mathcal{V}(\gamma_N)$, we have $Y_1 \subseteq \mathcal{V}(\gamma'_G)$, $\mathcal{V}(\gamma'_G) \cap Y_2 = \emptyset$, and $Y_2 \subseteq \mathcal{V}(\gamma''_G)$. We will show that

$$\beta' := \exists Y_1(\gamma_N \wedge \gamma'_G)$$

does the job.

It is straightforward to verify that $\beta'$ is a prime formula, and that $\mathcal{V}(\beta') \subseteq \mathcal{V}(\exists x \beta)$.

Next we show $\exists Y_2 \gamma''_G \models_{FT} \top$. Since $\gamma''_G$ is a solved clause and $Y_2$ contains all variables that are constrained in $\gamma''_G$, we know by the third axiom scheme that $FT \models \tilde{\forall} \exists Y_2 \gamma''_G$.

Finally we show $\exists x \beta \models_{FT} \beta'$. To see this, recall $\mathcal{V}(\gamma_N) \cap Y = \emptyset$ and $\mathcal{V}(\gamma'_G) \cap Y_2 = \emptyset$, and consider:

$$\begin{aligned}
\exists x \beta &= \exists x \exists X (\gamma_N \wedge \gamma_G) \\
&\models \exists Y (\gamma_N \wedge \gamma_G) \\
&\models \gamma_N \wedge \exists Y \gamma_G \\
&\models \gamma_N \wedge \exists Y_1 \exists Y_2 (\gamma'_G \wedge \gamma''_G) \\
&\models \gamma_N \wedge \exists Y_1 (\gamma'_G \wedge \exists Y_2 \gamma''_G) \\
&\models_{FT} \gamma_N \wedge \exists Y_1 \gamma'_G \\
&\models \exists Y_1 (\gamma_N \wedge \gamma'_G) = \beta'.
\end{aligned}$$

□

**Proposition 7.3** *If $\beta$ is a prime formula, then $FT \models \tilde{\exists} \beta$.*

**Proof.** Follows from the preceding proposition since $\top$ is the only closed prime formula. □

The next proposition establishes that prime formulae are closed under consistent conjunction (property (3) of Lemma 4.1).

**Proposition 7.4** *For every two prime formulae $\beta$ and $\beta'$ one can compute a formula $\delta$ that is either prime or $\bot$ and satisfies*

$$\beta \wedge \beta' \models_{FT} \delta \quad \text{and} \quad \mathcal{V}(\delta) \subseteq \mathcal{V}(\beta \wedge \beta').$$

**Proof.** Let $\beta = \exists X \gamma$ and $\beta' = \exists X' \gamma'$ be prime formulae. Without loss of generality we can assume that $X$ and $X'$ are disjoint. Hence

$$\beta \wedge \beta' \; \models\!\!\mid \; \exists X \exists X' (\gamma \wedge \gamma').$$

Since $\gamma \wedge \gamma'$ is a basic formula, Proposition 5.2 tells us that we can compute a formula $\phi$ that is either solved or $\bot$ and satisfies $\gamma \wedge \gamma' \models_{FT} \phi$ and $\mathcal{V}(\phi) \subseteq \mathcal{V}(\gamma \wedge \gamma')$. If $\phi = \bot$, then $\delta := \bot$ does the job. Otherwise, $\phi$ is solved. Since

$$\beta \wedge \beta' \; \models_{FT} \; \exists X \exists X' \phi,$$

we know by Proposition 7.2 how to compute a prime formula $\beta''$ such that $\beta \wedge \beta' \models_{FT} \beta''$. From the construction of $\beta''$ one verifies easily that $\mathcal{V}(\beta'') \subseteq \mathcal{V}(\beta \wedge \beta')$. $\square$

**Proposition 7.5** *Let $\phi$ be a formula that is built from atomic formulae with conjunction and existential quantification. Then one can compute a formula $\delta$ that is either prime or $\bot$ such that $\phi \models_{FT} \delta$ and $\mathcal{V}(\delta) \subseteq \mathcal{V}(\phi)$.*

**Proof.** Follows with Propositions 7.2 and 7.4. $\square$

The **closure of a prime formula** $\exists X \gamma$ is defined as follows:

$$[\exists X \gamma] \; := \; \{\pi \in [\gamma] \mid \mathcal{V}(\pi) \cap X = \emptyset \text{ or } \pi = x \varepsilon x \text{ or } \pi = x \varepsilon {\downarrow} x \varepsilon\}.$$

The **proper closure of a prime formula** $\beta$ is defined as follows:

$$[\beta]^\star \; := \; \{\pi \in [\beta] \mid \pi \text{ is a proper path constraint}\}.$$

**Proposition 7.6** *If $\beta$ is a prime formula and $\pi \in [\beta]$, then $\beta \models \pi$ (and hence $\neg \pi \models \neg \beta$).*

**Proof.** Let $\beta = \exists X \gamma$ be a prime formula, $\mathcal{A}, \alpha \models \beta$, and $\pi \in [\beta]$. Then there exists a $X$-update $\alpha'$ of $\alpha$ such that $\mathcal{A}, \alpha' \models \gamma$. Since $[\beta] \subseteq [\gamma]$, we have $\pi \in [\gamma]$ and thus $\mathcal{A}, \alpha' \models \pi$. If $\pi$ has no variable in common with $X$, then $\mathcal{A}, \alpha \models \pi$. Otherwise, $\pi$ has the form "$x \varepsilon x$" or "$x \varepsilon {\downarrow} x \varepsilon$" and hence $\mathcal{A}, \alpha \models \pi$ holds trivially. $\square$

We now know that the closure $[\beta]$, taken as an infinite conjunction, is entailed by $\beta$. We are going to show that, conversely, $\beta$ is entailed by certain finite subsets of its closure $[\beta]$.

An **access function** for a prime formula $\beta = \exists X \gamma$ is a function that maps every $x \in \mathcal{V}(\gamma) - X$ to the rooted path $x\varepsilon$, and every $x \in X$ to a rooted path $x'p$ such that $x'px \in [\gamma]$ and $x' \notin X$. Note that every prime formula has at least one access function, and that the access function of a prime formula is injective on $\mathcal{V}(\gamma)$ (follows from Proposition 5.3 (5)).

The **projection** of a prime formula $\beta = \exists X \gamma$ with respect to an access function @ for $\beta$ is the conjunction of the following proper path constraints:

$$\{x\varepsilon \downarrow y\varepsilon \mid x \doteq y \in \gamma\} \cup$$
$$\{Ax'p \mid Ax \in \gamma,\ x'p = @x\} \cup$$
$$\{x'pf \downarrow y'q \mid xfy \in \gamma,\ x'p = @x,\ y'q = @y\}.$$

Obviously, one can compute for every prime formula an access function and hence a projection. Furthermore, if $\lambda$ is a projection of a prime formula $\beta$, then $\lambda$ taken as a set is a finite subset of the closure $[\beta]$.

**Proposition 7.7** *Let $\lambda$ be a projection of a prime formula $\beta$. Then $\lambda \subseteq [\beta]^\star$ and $\lambda \models_{FT} \beta$.*

**Proof.** Let $\lambda$ be the projection of a prime formula $\beta = \exists X \gamma$ with respect to an access function @.

Since every path constraint $\pi \in \lambda$ is in $[\beta]$ and thus satisfies $\beta \models \pi$, we have $\beta \models \lambda$.

To show the other direction, suppose $\mathcal{A}, \alpha \models \lambda$, where $\mathcal{A}$ is a model of $FT$. Then $\mathcal{A}, \alpha' \models x'px$ for every $x \in X$ with $@x = x'p$ defines a unique $X$-update $\alpha'$ of $\alpha$. From the definition of a projection it is clear that $\mathcal{A}, \alpha' \models \gamma$. Hence $\mathcal{A}, \alpha \models \beta$. □

As a consequence of this proposition one can compute for every prime formula an equivalent quantifier-free conjunction of proper path constraints.

We close this section with a few propositions stating interesting properties of closures of prime formulae. These propositions will not be used in the proofs to come. The reader is nevertheless advised to study the proof of Proposition 7.9 since it employs a construction that will be reused in a more complicated form in the proof of Lemma 8.4.

**Proposition 7.8** *If $\beta$ is a prime formula, then $\beta \models_{FT} [\beta]^\star$.*

**Proof.** By Proposition 7.6 we have $\beta \models_{FT} [\beta]^\star$, and by Proposition 7.7 we have $[\beta]^\star \models_{FT} \beta$ since $\beta$ has a projection $\lambda \subseteq [\beta]^\star$. □

**Proposition 7.9** *If $\beta$ is a prime formula, and $\pi$ is a proper path constraint, then*

$$\pi \in [\beta]^\star \iff \beta \models_{FT} \pi.$$

**Proof.** Let $\beta = \exists X \gamma$ be a prime formula, $\gamma = \gamma_N \wedge \gamma_G$ be the decomposition of $\gamma$ into graph and normalizer, and $\pi$ be a proper path constraint. Since the direction "⇒" is stated by Proposition 7.6, it suffices to show the other direction.

Suppose $\pi \notin [\beta]$. We show that $FT \models \tilde{\exists}(\beta \wedge \neg\pi)$, which yields $\beta \not\models_{FT} \pi$ since $FT$ is consistent.

Without loss of generality we can assume that $\mathcal{V}(\pi)$ and $X$ are disjoint. Let $Y$ be the variables eliminated by $\gamma$. Since $(\beta \wedge \neg\pi) \models\!\!\!\mid (\beta \wedge \neg\pi[x \leftarrow y])$ if $x \doteq y \in \gamma_N$, we can assume without loss of generality that $\pi$ contains no variable in $Y$.

Since

$$\begin{aligned}
\tilde{\exists}(\beta \wedge \neg\pi) &\models\!\!\!\mid \tilde{\exists}\exists Y(\gamma_N \wedge \exists X \gamma_G \wedge \neg\pi) \\
&\models\!\!\!\mid \tilde{\exists}(\exists Y \gamma_N \wedge \exists X \gamma_G \wedge \neg\pi) \\
&\models\!\!\!\mid \tilde{\exists}(\exists X \gamma_G \wedge \neg\pi) \\
&\models\!\!\!\mid \tilde{\exists}(\gamma_G \wedge \neg\pi),
\end{aligned}$$

it is sufficient to construct a solved clause $\delta$ with $\gamma_G \subseteq \delta$ and $\delta \models_{FT} \neg\pi$ (recall that $FT \models \tilde{\exists}\delta$ by the third axiom scheme). For the construction of $\delta$ we distinguish three cases:

1. $\pi = Axp$, $\pi = xp \downarrow yq$ or $\pi = yq \downarrow xp$, where $xp \downarrow xp \notin [\gamma_G]$. Then there exists a prefix $p'f$ of $p$ and a variable $z$ such that $xp'z \in [\gamma_G]$ and $zfz' \in \gamma_G$ for no variable $z'$. Now adding $zf\uparrow$ yields a solved clause $\delta$ such that $\delta \models_{FT} \neg\pi$.

2. $\pi = Axp$, $xpz \in [\gamma_G]$. If $Bz \in \gamma_G$, then $A \neq B$ (since $\pi \notin [\gamma_G]$) and $\delta := \gamma_G$ does the job. Otherwise, we choose a sort $B \neq A$ and add $Bz$ (recall that we have assumed infinitely many sorts).

3. $\pi = xp \downarrow yq$, $xpz \in [\gamma_G]$ and $yqz' \in [\gamma_G]$. Since $\pi \notin [\beta]$, we know that $z \neq z'$. We choose a new feature $f$ and a new variable $u$ and add $zf\uparrow$ and $z'fu$ (recall that we have assumed infinitely many features). □

**Proposition 7.10** *Let $\beta$, $\beta'$ be prime formulae. Then*

$$\beta \models_{FT} \beta' \iff [\beta]^\star \supseteq [\beta']^\star.$$

**Proof.** "$\Rightarrow$" Let $\beta \models_{FT} \beta'$ and $\pi \in [\beta']^\star$. Then $\beta' \models_{FT} \pi$ by Proposition 7.6 and hence $\beta \models_{FT} \pi$ by the assumption. Hence $\pi \in [\beta]^\star$ by Proposition 7.9.

"$\Leftarrow$" Let $[\beta]^\star \supseteq [\beta']^\star$. Then $[\beta]^\star \models [\beta']^\star$ and hence $\beta \models_{FT} \beta'$ by Proposition 7.8. □

**Proposition 7.11** *Let $\beta$, $\beta'$ be prime formulae, and let $\lambda'$ be a projection of $\beta'$. Then $\beta \models_{FT} \beta' \iff [\beta]^\star \supseteq \lambda'$.*

**Proof.** "$\Rightarrow$" Suppose $\beta \models_{FT} \beta'$. Then $[\beta]^\star \supseteq [\beta']^\star$ by Proposition 7.10 and $[\beta]^\star \supseteq \lambda'$ by Proposition 7.7.

"$\Leftarrow$" Suppose $[\beta]^\star \supseteq \lambda'$. Then $[\beta]^\star \models \lambda'$ and hence $\beta \models_{FT} \beta'$ by Proposition 7.8 and 7.7. □

Proposition 7.11 gives us a decision procedure for "$\beta \models_{FT} \beta'$" since membership in $[\beta]^\star$ is decidable, $\lambda'$ is finite, and $\lambda'$ can be computed from $\beta'$.

## 8 Quantifier Elimination

In this section we show that our prime formulae satisfy the requirements (5) and (6) of Lemma 4.1 and thus obtain the completeness of *FT*. We start with the definition of the central notion of a joker.

A **rooted path** $xp$ consists of a variable $x$ and a path $p$. A rooted path $xp$ is called **unfree** in a prime formula $\beta$ if

$$\exists \text{ prefix } p' \text{ of } p \; \exists \, yq: \quad x \neq y \text{ and } xp'{\downarrow}yq \in [\beta].$$

A rooted path is called **free** in a prime formula $\beta$ if it is not unfree in $\beta$.

**Proposition 8.1** *Let $\beta = \exists X \gamma$ be a prime formula. Then:*

1. *if $xp$ is free in $\beta$, then $x$ does not occur in the normalizer of $\gamma$*

2. *if $x \notin \mathcal{V}(\beta)$, then $xp$ is free in $\beta$ for every path $p$.*

A proper path constraint $\pi$ is called an $x$-**joker** for a prime formula $\beta$ if $\pi \notin [\beta]$ and one of the following conditions is satisfied:

1. $\pi = Axp$ and $xp$ is free in $\beta$

2. $\pi = xp{\downarrow}yq$ and $xp$ is free in $\beta$

3. $\pi = yp{\downarrow}xq$ and $xq$ is free in $\beta$.

**Proposition 8.2** *It is decidable whether a rooted path is free in a prime formula, and whether a path constraint is an $x$-joker for a prime formula.*

**Proof.** Follows with Proposition 5.3. □

**Lemma 8.3** *Let $\beta$ be a prime formula, $x$ be a variable, $\pi$ be a proper path constraint that is not an $x$-joker for $\beta$, $\mathcal{A}$ be a model of $FT$, $\mathcal{A}, \alpha \models \beta$, $\mathcal{A}, \alpha' \models \beta$, and $\alpha'$ be an $x$-update of $\alpha$. Then $\mathcal{A}, \alpha \models \pi$ if and only if $\mathcal{A}, \alpha' \models \pi$.*

**Proof.** We distinguish the following cases:

1. $x \notin \mathcal{V}(\pi)$. Then the claim is trivial.

2. $\pi \in [\beta]$. Then $\beta \models_{FT} \pi$ and hence $\alpha, \alpha' \in \pi^{\mathcal{A}}$.

3. $\pi = Axp$ and $xp$ unfree in $\beta$. Then $p = p'p''$ and $xp'{\downarrow}yq \in [\beta]$ for some variable $y \neq x$ and some path $q$. Hence $\beta \models_{FT} \pi \leftrightarrow Ayqp''$, which yields the claim.

4. $\pi = xp{\downarrow}yq$, $x \neq y$, $xp$ unfree in $\beta$. Analogous to case (3).

5. $\pi = xp{\downarrow}xq$ and both $xp$, $xq$ unfree in $\beta$. Analogous to case (3). □

**Lemma 8.4** *Let $\beta$ be a prime formula and $\pi_1, \ldots, \pi_n$ be $x$-jokers for $\beta$. Then*

$$\exists x \beta \models_{FT} \exists x (\beta \wedge \bigwedge_{i=1}^{n} \neg \pi_i).$$

**Proof.** Let $\beta = \exists X \gamma$ be a prime formula, $\pi_1, \ldots, \pi_n$ ($n > 0$) be $x$-jokers for $\beta$, $\mathcal{A}$ be a model of $FT$, and $\alpha$ be a valuation into $\mathcal{A}$ such that $\mathcal{A}, \alpha \models \exists x \beta$. We have to show that $\mathcal{A}, \alpha \models \exists x (\beta \wedge \bigwedge_{i=1}^{n} \neg \pi_i)$. Without loss of generality we assume that $x \notin X$, and that no $\pi_i$ has a variable in common with $X$. Let $\gamma = \gamma_N \wedge \gamma_G$ be the decomposition of $\gamma$ into normalizer and graph. Since there are $x$-jokers for $\beta$, we know that $x \notin \mathcal{V}(\gamma_N)$.

The proof now comes in two parts. Part II gives the construction of a solved clause $\delta$ such that, if $Y$ and $Y_1$ are defined as

$$Y := \{x\} \cup X \cup (\mathcal{V}(\delta) - \mathcal{V}(\gamma_G)) \quad \text{"quantified variables"}$$
$$Y_1 := \{y \in Y \mid \forall y'py \in [\delta] : y' \in Y\} \quad \text{"unreachable variables"},$$

the following conditions are satisfied:

1. $\gamma_G \subseteq \delta$

2. additional variables in $\delta$ are new variables, that is, $(\mathcal{V}(\delta) - \mathcal{V}(\gamma_G)) \cap \mathcal{V}(\gamma_N) = \emptyset$ and $(\mathcal{V}(\delta) - \mathcal{V}(\gamma_G)) \cap \mathcal{V}(\pi_i) = \emptyset$ for $i = 1, \ldots, n$

3. if $\alpha'$ is an $Y$-update of $\alpha$ such that $\mathcal{A}, \alpha' \models \delta$, then $\mathcal{A}, \alpha' \models \neg \pi_i$ for $i = 1, \ldots, n$

4. every atomic formula that occurs in $\delta$ but not in $\gamma_G$ contains only variables in $Y_1$.

In Part I of the proof we will show that from the existence of a solved clause $\delta$ as specified above we can derive $\mathcal{A}, \alpha \models \exists x (\beta \wedge \bigwedge_{i=1}^{n} \neg \pi_i)$. Part I uses a garbage collection technique similar to the one used in the proof of Proposition 7.2. The construction of $\delta$ in Part II is a refinement of the construction in the proof of Proposition 7.9. We strongly recommend that the reader first gets a good intuitive understanding of the proofs of Proposition 7.2 and 7.9 before studying the rest of this proof.

*Part I.* Suppose $\delta$, $Y$ and $Y_1$ are given as specified above. We define $Y_2$, $\delta_1$ and $\delta_2$ such that

- $Y = Y_1 \uplus Y_2$
- $\delta = \delta_1 \wedge \delta_2$
- $\mathcal{V}(\delta_2) \cap Y_1 = \emptyset$
- every atomic formula in $\delta_1$ contains a variable in $Y_1$.

To stay with the garbage collection metaphor, think of $Y_2$ as the reachable variables, of $\delta_1$ as the unreachable part of $\delta$, and $\delta_2$ as the reachable part of $\delta$. By assumption (4) we know that $\delta_2 \subseteq \gamma_G$. By the third axiom scheme we know that $\exists Y_1 \delta_1 \models_{FT} \top$, since $\delta_1$ is a solved clause and $Y_1$ contains all variables that are constrained in $\delta_1$.

Note that $\{x\}$, $X$ and $\mathcal{V}(\delta) - \mathcal{V}(\gamma_G)$ are pairwise disjoint. Hence

$$\exists x \beta \models_{FT} \gamma_N \wedge \exists Y \delta$$

since

$$\exists x \beta \mathrel{|\!\!=\!\!|} \exists x \exists X (\gamma_N \wedge \gamma_G) \mathrel{|\!\!=\!\!|} \gamma_N \wedge \exists x \exists X \gamma_G \mathrel{|\!\!=\!\!|} \gamma_N \wedge \exists Y \gamma_G$$

and

$$\exists Y \gamma_G \models \exists Y \delta_2 \mathrel{|\!\!=\!\!|}_{FT} \exists Y (\delta_2 \wedge \exists Y_1 \delta_1) \mathrel{|\!\!=\!\!|}_{FT} \exists Y (\delta_2 \wedge \delta_1) \mathrel{|\!\!=\!\!|}_{FT} \exists Y \delta.$$

Thus $\mathcal{A}, \alpha \models \gamma_N \wedge \exists Y \delta$. Since $\mathcal{V}(\gamma_N) \cap Y = \emptyset$, there exists an $Y$-update of $\alpha'$ such that $\mathcal{A}, \alpha' \models \gamma_N \wedge \delta$. By assumption (3) we know that $\mathcal{A}, \alpha' \models \neg \pi_i$ for $i = 1, \ldots, n$, and by assumption (1) we know that $\mathcal{A}, \alpha' \models \gamma_G$. Thus $\mathcal{A}, \alpha' \models \exists Y (\gamma \wedge \bigwedge_{i=1}^{n} \neg \pi_i)$. Since $\mathcal{V}(\delta) - \mathcal{V}(\gamma_G)$ has no variable in common with $\gamma \wedge \bigwedge_{i=1}^{n} \neg \pi_i$ and $X$ has no variable in common with $\bigwedge_{i=1}^{n} \neg \pi_i$, we have $\mathcal{A}, \alpha' \models \exists x (\beta \wedge \bigwedge_{i=1}^{n} \neg \pi_i)$.

*Part II.* We will now construct a solved form $\delta$ as required. To do this we will look at every $x$-joker $\pi_i$ and possibly add constraints to $\gamma_G$ such that requirement (3) in particular is satisfied. It suffices to distinguish the following cases (recall that $x \notin \mathcal{V}(\gamma_N)$):

1. $\pi_i = Axp$, $xpz \in [\gamma_G]$. If $Bz \in \gamma_G$, then $A \neq B$ (since $\pi_i \notin [\gamma_G]$) and requirement (3) is met without adding anything. Otherwise, we choose a new sort $B$ and add $Bz$ (recall that we have assumed infinitely many sorts).

2. $\pi_i = Axp$, $xp \downarrow xp \notin [\gamma_G]$. Then there exists a prefix $p'f$ of $p$ and a variable $z$ such that $xp'z \in [\gamma_G]$ and $zfz' \notin \gamma_G$ for every $z'$. Adding $zf\uparrow$ will yield a solved form and satisfy the requirements (1)–(3). It will also satisfy requirement (4) since $xp$ is free in $\beta$.

3. $\pi_i = xp \downarrow yq$, $xp$ free in $\beta$, $xp \downarrow xp \notin [\gamma_G]$. Analogous to case (2).

4. $\pi_i = xp \downarrow yq$, $xp$ free in $\beta$, $xpz \in [\gamma_G]$. We once more distinguish three cases:

   4.1 $x \neq y$. Let $\alpha'$ be a $Y$-update of $\alpha$ such that $\mathcal{A}, \alpha' \models \gamma$. Then $q^{\mathcal{A}}$ is defined on $\alpha'(y)$ if and only if $q^{\mathcal{A}}$ is defined on $\alpha(y)$. If $q^{\mathcal{A}}$ is undefined on $\alpha(y)$, requirement (3) is satisfied without adding anything. Otherwise, let $\alpha(y) q^{\mathcal{A}} a$. Then $\alpha'(y) q^{\mathcal{A}} a$. Now choose a new feature $f$ (recall that we have infinitely many features). If $f^{\mathcal{A}}$ is defined on $a$, we add $zf\uparrow$; otherwise we add $zfz'$, were $z'$ is a new variable. Requirements (1)–(3) are obviously satisfied, and requirement (4) is satisfied since $xp$ is free in $\beta$.

   4.2 $x = y$ and $xq$ unfree in $\beta$. Then we have $q = q'q''$, $xq' \downarrow y'r \in [\beta]$ and $y' \notin Y$ for some $q'$, $q''$ $y'$ and $r$. Let $\alpha'$ be a $Y$-update of $\alpha$ such that $\mathcal{A}, \alpha' \models \gamma$. Then $q^{\mathcal{A}} = q'^{\mathcal{A}} q''^{\mathcal{A}}$ is defined on $\alpha'(x)$ if and only if $r^{\mathcal{A}} q''^{\mathcal{A}}$ is defined on $\alpha(y')$. If $r^{\mathcal{A}} q''^{\mathcal{A}}$ is undefined on $\alpha(y')$, requirement (3) is satisfied without adding anything. Otherwise, let $\alpha(y') r^{\mathcal{A}} q''^{\mathcal{A}} a$. Then $\alpha'(x) q^{\mathcal{A}} a$. Now choose a new feature $f$. If $f^{\mathcal{A}}$ is defined on $a$, add $zf\uparrow$; otherwise, add $zfz'$, where $z'$ is a new variable. Requirements (1)–(3) are obviously satisfied, and requirement (4) is satisfied since $xp$ is free in $\beta$.

4.3 $x = y$ and $xq$ free in $\beta$. If $xq \downarrow xq \notin [\gamma_G]$, we proceed analogous to case (2). Otherwise, let $xqz' \in [\gamma_G]$. Since $\pi_i \notin [\beta]$, we know that $z \neq z'$. We choose a new feature $f$ and a new variable $u$ and add $zf\uparrow$ and $z'fu$. This will certainly satisfy the requirements (1)–(3). It will also satisfy requirement (4) since both $xp$ and $xq$ are free in $\beta$.

$\square$

Note that the proof uses the third axiom scheme, the existence of infinitely many features, and the existence of infinitely many sorts.

**Lemma 8.5** *Let $\beta$, $\beta'$ be prime formulae and $\alpha$ be a valuation into a model $\mathcal{A}$ of $FT$ such that*

$$\mathcal{A}, \alpha \models \exists x (\beta \wedge \beta') \quad \text{and} \quad \mathcal{A}, \alpha \models \exists x (\beta \wedge \neg \beta').$$

*Then every projection of $\beta'$ contains an $x$-joker for $\beta$.*

**Proof.** Without loss of generality we can assume that $\mathcal{A}, \alpha \models \beta \wedge \beta'$. Furthermore, there exists an $x$-update $\alpha'$ of $\alpha$ such that $\mathcal{A}, \alpha' \models \beta \wedge \neg \beta'$. Let $\lambda$ be a projection of $\beta'$. Since $\mathcal{A}, \alpha' \not\models \beta'$, we know by Proposition 7.7 that $\mathcal{A}, \alpha' \not\models \lambda$. Hence there exists a proper path constraint $\pi \in \lambda$ such that $\mathcal{A}, \alpha' \not\models \pi$. Since $\mathcal{A}, \alpha \models \beta'$, we know by Proposition 7.6 that $\mathcal{A}, \alpha \models \pi$. Hence we know by Lemma 8.3 that $\pi$ must be an $x$-joker for $\beta$. $\square$

**Lemma 8.6** *If $\beta, \beta_1, \ldots, \beta_n$ are prime formulae, then*

$$\exists x (\beta \wedge \bigwedge_{i=1}^{n} \neg \beta_i) \quad \dashv\vdash_{FT} \quad \bigwedge_{i=1}^{n} \exists x (\beta \wedge \neg \beta_i).$$

**Proof.** Let $\beta, \beta_1, \ldots, \beta_n$ be prime formulae. Then $\exists x (\beta \wedge \bigwedge_{i=1}^{n} \neg \beta_i) \models \bigwedge_{i=1}^{n} \exists x (\beta \wedge \neg \beta_i)$ is trivial. To see the other direction, suppose that $\mathcal{A}$ is a model of $FT$ and $\mathcal{A}, \alpha \models \bigwedge_{i=1}^{n} \exists x (\beta \wedge \neg \beta_i)$. We have to exhibit some $x$-update $\alpha'$ of $\alpha$ such that $\mathcal{A}, \alpha' \models \beta$ and $\mathcal{A}, \alpha' \models \neg \beta_i$ for $i = 1, \ldots, n$.

Without loss of generality we can assume that $\mathcal{A}, \alpha' \models \exists x (\beta \wedge \beta_i)$ for $i = 1, \ldots, m$ and $\mathcal{A}, \alpha' \models \neg \exists x (\beta \wedge \beta_i)$ for $i = m+1, \ldots, n$.

By Lemma 8.5 there exists, for every $i = 1, \ldots, m$, an $x$-joker $\pi_i \in [\beta_i]$ for $\beta$. By Lemma 8.4 we have

$$\exists x \beta \models \exists x (\beta \wedge \bigwedge_{i=1}^{m} \neg \pi_i).$$

Since $\neg \pi \models \neg \beta_i$ by Proposition 7.6, we have

$$\exists x \beta \models \exists x (\beta \wedge \bigwedge_{i=1}^{m} \neg \beta_i).$$

Hence we know that there exists an $x$-update $\alpha'$ of $\alpha$ such that $\mathcal{A}, \alpha' \models \beta$ and $\mathcal{A}, \alpha' \models \neg \beta_i$ for $i = 1, \ldots, m$. Since we know that $\mathcal{A}, \alpha \models \neg \exists x (\beta \wedge \beta_i)$ for $i = m+1, \ldots, n$, we have $\mathcal{A}, \alpha' \models \neg \beta_i$ for $i = m+1, \ldots, n$. $\square$

**Lemma 8.7** *For every two prime formulae $\beta$, $\beta'$ and every variable $x$ one can compute a Boolean combination $\delta$ of prime formulae such that*

$$\exists x (\beta \wedge \neg \beta') \models\!\!\!\models_{FT} \delta \quad \text{and} \quad \mathcal{V}(\delta) \subseteq \mathcal{V}(\exists x (\beta \wedge \neg \beta')).$$

**Proof.** Let $\beta, \beta'$ be prime formulae, $\lambda$ be a projection of $\beta'$, $x$ be a variable and $\mathcal{A}$ be a model of $FT$. We distinguish two cases:

1. *$\lambda$ contains an $x$-joker $\pi$ for $\beta$.* Then we know that $\exists x \beta \models \exists x (\beta \wedge \neg \pi)$ by Lemma 8.4. Since $\beta' \models_{FT} \lambda \models \pi$, we know that $\neg \pi \models \neg \beta'$ and hence $\exists x \beta \models_{FT} \exists x (\beta \wedge \neg \beta')$. Thus

$$\exists x (\beta \wedge \neg \beta') \models\!\!\!\models_{FT} \exists x \beta.$$

Now the claim follows with Proposition 7.2.

2. *$\lambda$ contains no $x$-joker $\pi$ for $\beta$.* Then we know by Lemma 8.5 that there exists no valuation $\alpha$ into $\mathcal{A}$ such that

$$\mathcal{A}, \alpha \models \exists x (\beta \wedge \beta') \quad \text{and} \quad \mathcal{A}, \alpha \models \exists x (\beta \wedge \neg \beta').$$

Hence
$$\exists x (\beta \wedge \neg \beta') \models\!\!\!\models_{FT} \exists x \beta \wedge \neg \exists x (\beta \wedge \beta').$$

Now the claim follows with Propositions 7.2, 7.4 and 8.2.

The above shows the existence of $\delta$. Moreover, $\delta$ can be computed since we can compute a projection $\lambda$ of $\beta'$, and since we can decide whether $\lambda$ contains an $x$-joker for $\beta$ by Proposition 8.2 ($\lambda$ is finite). $\square$

**Theorem 8.8** *For every formula $\phi$ one can compute a Boolean combination $\delta$ of prime formulae such that $\phi \models\!\!\!\models_{FT} \delta$ and $\mathcal{V}(\delta) \subseteq \mathcal{V}(\beta).$*

**Proof.** Follows from Lemma 4.1, Propositions 7.4 and 7.2, and Lemmas 8.6 and 8.7. □

**Corollary 8.9** *FT is a complete and decidable theory.*

**Proof.** The completeness of *FT* follows from the preceding theorem and the fact that ⊤ is the only closed prime formula. The decidability follows from the completeness and the fact that *FT* is given by a recursive set of sentences. □

# Acknowledgements

We appreciate discussions with Joachim Niehren and Ralf Treinen who read a draft version of this paper.

The research reported in this paper has been supported by the Bundesminister für Forschung und Technologie under contracts ITW 90002 0 (DISCO) and ITW 9105 (Hydra).